\pdfoutput=1
\documentclass[11pt]{article}

\usepackage[final]{acl}

\usepackage{times}
\usepackage{latexsym}
\usepackage{tabularx}
\usepackage{amsmath,amssymb}
\usepackage{multirow}
\usepackage{booktabs}
\usepackage{multirow}
\usepackage{adjustbox}
\usepackage{amssymb}
\usepackage{wasysym}
\usepackage{caption}
\usepackage{subcaption}
\usepackage{xcolor}
\usepackage{makecell}

\definecolor{mygray}{gray}{.9}
\usepackage{pifont}% http://ctan.org/pkg/pifont
\newcommand{\cmark}{\ding{51}}%
\newcommand{\xmark}{\ding{55}}%

\usepackage[T1]{fontenc}
\usepackage[utf8]{inputenc}

\usepackage{microtype}

\usepackage{inconsolata}

\usepackage{graphicx}

\title{Dynamic Fisher-weighted Model Merging via Bayesian Optimization}

\author{Sanwoo Lee$^{1}$\thanks{~~work done during an internship at Meituan.} ,
Jiahao Liu$^{2}$,
Qifan Wang$^{3}$,
Jingang Wang$^{2}$,
Xunliang Cai$^{2}$,
Yunfang Wu$^{1}$\thanks{~~Corresponding author.} \\
$^{1}$School of Computer Science, Peking University; $^{2}$Meituan; $^{3}$Meta AI \\
\texttt{\{sanwoo, wuyf\}@pku.edu.cn}, \texttt{wqfcr@fb.com} \\ \texttt{\{liujiahao12,wangjingang02,caixunliang\}@meituan.com}}

\begin{document}
\maketitle
\begin{abstract}
The fine-tuning of pre-trained language models has resulted in the widespread availability of task-specific models. Model merging offers an efficient way to create multi-task models by combining these fine-tuned models at the parameter level, without the need for training data or joint training on multiple datasets. Existing merging approaches typically involve scaling the parameters model-wise or integrating parameter importance parameter-wise. Both approaches exhibit their own weaknesses, leading to a notable performance gap compared to multi-task fine-tuning.
In this paper, we unify these seemingly distinct strategies into a more general merging framework, and introduce \textbf{D}ynamic \textbf{F}isher-weighted \textbf{M}erging (\textbf{DF-Merge})\footnote{Code is available at \url{https://github.com/sanwooo/df-merge}}. Specifically, candidate models are associated with a set of coefficients that linearly scale their fine-tuned parameters. Bayesian optimization is applied to dynamically adjust these coefficients, aiming to maximize overall performance on validation sets. Each iteration of this process integrates parameter importance based on the Fisher information conditioned by the coefficients. 
Experimental results show that DF-Merge outperforms strong baselines across models of different sizes and a variety of tasks. Our analysis shows that the effectiveness of DF-Merge arises from the unified view of merging and that near-optimal performance is achievable in a few iterations, even with minimal validation data.

\end{abstract}

\section{Introduction}
Modern transformer-based pre-trained language models (PLMs) \citep{devlin-etal-2019-bert, raffel2020exploring, NEURIPS2020_1457c0d6} have driven a paradigm shift towards fine-tuning PLMs for specific tasks, achieving state-of-the-art performance across various applications. The general-purpose representations learned through pretraining have significantly enhanced numerous downstream tasks, leading to the widespread development of fine-tuned expert models \citep{min2023recent}. For example, over a million models have been uploaded to the Hugging Face repository \citep{wolf2019huggingface}, with many publicly available for research study\footnote{https://huggingface.co/models}.

\begin{figure}[t]
    \centering
    \includegraphics[width=1\columnwidth]{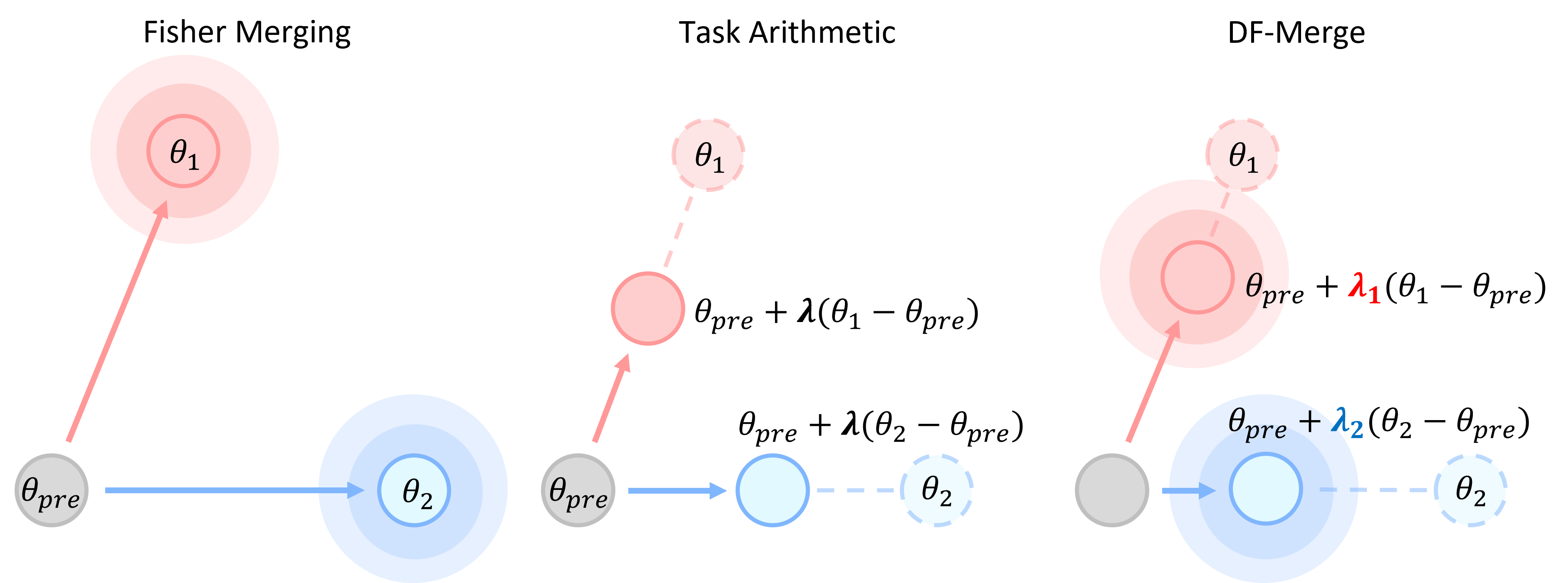}
    \caption{ Comparison of DF-Merge with primary approaches in model merging. \textbf{left}: leverages parameter importance evaluated at the fine-tuned models. \textbf{middle}: uniformly scales fine-tuned models to alleviate parameter interference. \textbf{right}: DF-Merge optimizes distinct scaling coefficients and incorporates parameter importance evaluated at the scaled models.}
    \label{fig:overview}
\end{figure}

Most off-the-shelf models are fine-tuned independently for individual tasks, which limits their performance outside of their specialized domains. Ideally, models should be capable of handling multiple tasks relevant to a particular use case. Although multi-task learning (MTL) \cite{sogaard-goldberg-2016-deep, deng2019multi} offers a straightforward solution, it must require simultaneous access to the labeled datasets of all tasks and training over those datasets. This challenge is pronounced given the increasing difficulty of fine-tuning PLMs with their ever-growing sizes. More importantly, since the original training data for each model is often proprietary, it is time-consuming or even infeasible for users to label a large amount of data for MTL.

Model merging offers a cost-effective alternative for building multi-task models by combining off-the-shelf models in the parameter space without additional training. For example, this can be done by simply weight-averaging the \textit{task vectors} (i.e. fine-tuned part of the parameters from the pre-trained model) \citep{ilharco2023editing, yang2024adamerging}. The success of model merging is supported by recent findings that the local minima, optimized from pre-trained parameters, are linearly connected in a flat basin of the loss landscape with no barriers in between \citep{neyshabur2020being, zhou2024on}. As a result, linearly interpolating between fine-tuned models potentially produces a well-behaved model with multi-task capabilities.

Despite the advantages of model merging, current methods still lag behind the performance of multi-task fine-tuned models. This shortfall can be attributed to the fact that existing approaches improve only specific aspects of model merging. These model merging methods can be divided into two groups: (1) scaling the task vectors \textit{model-wise} \cite{ilharco2023editing, yang2024adamerging, liu2024checkpoint}, and (2) accounting for parameter importance \textit{parameter-wise} \cite{matena2022merging, jin2023dataless, tam2024merging}. We present a general merging framework that these two seemingly distinct approaches can be unified into. Building on this framework, we introduce \textbf{D}ynamic \textbf{F}isher-weighted \textbf{M}erging (\textbf{DF-Merge}) which leverages the strengths of both strategies, as illustrated in Figure~\ref{fig:overview}. In essence, DF-Merge uses Bayesian optimization to adjust the scaling coefficients in order to maximize the overall performance, with each iteration targeting a low-loss basin informed by (approximated) Fisher information.

Experimental results show that DF-Merge significantly outperforms competitive baselines across PLMs of different sizes on a variety of tasks. Ablation study confirms that the components of DF-Merge collectively contributes to the performance, validating the advantage of the general merging framework. Additionally, our analysis demonstrates that DF-Merge can achieve near-optimal performance within just a few iterations using minimal validation data.
Our contributions are summarized as follows:
\begin{itemize}
    \item We formulate the two primary model merging approaches into an unified objective, achieving a more flexible and effective model merging framework.
    \item We introduce Bayesian optimization in model merging to identify the optimal coefficients which allows for direct maximization of non-differentiable metrics. 
    \item Our DF-Merge approach achieves significant improvements over the baselines, making it an effective and efficient alternative over multi-task learning.
\end{itemize}

\section{Model Merging Revisit}
\label{sec:background}

\paragraph{Notation.} Let $net(\theta)$ be a neural network parameterized by $\theta \in \mathbb{R}^{d}$. Consider $T$ task-specific models $\{ net(\theta_i) \}_{i=1}^{T}$, each initialized from the same pre-trained model $net(\theta_{pre})$ and fine-tuned on the $i$-th task dataset $\mathcal{D}_{i} = \{ x_{i}^{(j)}, y_{i}^{(j)} \}_{j=1}^{N_i}$ where $N_i$ is the dataset's cardinality. The goal of model merging is to create a multi-task model $net(\theta^{*})$ that is proficient in all tasks.

\paragraph{Task Arithmetic (TA).} \citet{ilharco2023editing} coined a concept of \textit{task vector} which represents the direction in the parameter space that enhances a pre-trained model's performance on the task. In particular, the task vector $\tau_{i}$ for the $i$-th task is specified by the fine-tuned part of parameters from the pre-trained model, expressed as $\tau_{i}=\theta_{i}-\theta_{pre}$. Task vectors can be combined by arithmetic operations to steer the pre-trained model's behavior on various tasks. This concept has been extended to \textit{model-wise} model merging \citep{yadav2024ties, yu2024language, yang2024adamerging}, in which multiple task vectors are added to the pre-trained parameters

\begin{equation}
    \theta_{new} = \theta_{pre} + \lambda\sum_{i=1}^{M} \Phi(\tau_{i})
\end{equation}

where $\lambda \in \mathbb{R}$ is a scaling coefficient and $\Phi$ denotes additional operations on the task vectors such as trimming, electing \citep{yadav2024ties}, or dropout \citep{yu2024language}. These operations are designed to reduce parameter interference across the fine-tuned models. While the original implementation optimizes for a single coefficient $\lambda$ on held-out validations sets, this can be generalized to multiple coefficients $\theta_{new} = \theta_{pre} + \sum_{i=1}^{M} \lambda_{i}\Phi(\tau_{i})$, which we name as \textbf{General Task Arithmetic} (\textbf{GTA}).

\paragraph{Fisher Information.} 

Understanding the landscape of the loss function allows for better alignment of different models' parameters. The local curvature of a loss function $\ell(\theta)$ at the point $\theta$ is captured by its second-order derivatives $\nabla^2 \ell(\theta)$, denoted as the Hessian matrix $H_{\theta} \in \mathbb{R}^{d \times d}$. Then the expectation of Hessian over the data distribution $p_{\theta}(x, y)$ describes how sensitive the loss function is to the parameters in the data distribution modeled by $\theta$, where highly sensitive parameters potentially imply greater importance.

Assuming the model is fine-tuned using negative log-likelihood loss $\ell(\theta) = -\log p(y|x, \theta)$, the expectation of $H_\theta$ can be efficiently computed by the Fisher information (FI):
\begin{equation}
    F_{\theta} = \mathop{\mathbb{E}}_{x \sim q(x) } \left[\mathop{\mathbb{E}}_{y\sim p_{\theta}(y|x)}\nabla_{\theta} \ell(\theta) \nabla_{\theta} \ell(\theta)^\top\right]
\end{equation}
which only requires computing the first-order derivatives.

As estimating the expectation over the input distribution  $x \sim q(x)$ is intractable, $F_\theta$ is approximated with the empirical Fisher information $\hat{F}_{\theta}$
\begin{equation}
    \hat{F}_{\theta} = \frac{1}{N}\sum_{i=1}^{N}\left[\mathop{\mathbb{E}}_{y\sim p_{\theta}(y|x^{(i)})}\nabla_{\theta} \ell(\theta) \nabla_{\theta} \ell(\theta)^\top\right]
\end{equation}

Note that the expectation over $y$ is not calculated over the true labels, but rather measured on the predictive distribution $y \sim p_{\theta}(y|x)$ parameterized by $\theta$. In practice, $y \sim p_{\theta}(y|x)$ can either be modeled exactly or through sampling depending on the size of label space \citep{matena2022merging}. 

\paragraph{Fisher Merging from Geometric Perspective.}
\citet{tam2024merging} studied a geometric analysis of Fisher Merging \citep{matena2022merging}, by representing Fisher Merging (without some approximations) as
\begin{equation}
 \small
    \theta^{*} = \left( \sum_{i=1}^{\mathrm{M}} Q_i \Lambda_i Q_i^{\top} \right)^{-1} \left( \sum_{i=1}^{\mathrm{M}} Q_i \Lambda_i Q_i^{\top} \theta_{i} \right)
\end{equation}
where $Q_i \Lambda_i Q_i^{\top}$ is the eigendecomposition of $F_{\theta_i}$. Inspecting this form, $Q_i \Lambda_i Q_i^{\top}$ upweights the "important" eigenvector component of $\theta_i$, such that useful parameters are preserved during merging. 

Building upon this insight, we consider the following geometric objective $g(\theta)$ and show that Fisher Merging is a natural result of minimizing it:
\begin{equation}
    \theta^{*} = \mathop{\arg \min}_{\theta} \sum_{i=1}^{\mathrm{M}} \| \Lambda_{i}^{1/2} (Q_{i}^{\top} \theta_{i} - Q_{i}^{\top} \theta) \|^{2}
\end{equation}
which restricts $\theta$ to move along the loss-insensitive principal directions in the parameter space, as indicated by the eigenvectors associated with smaller eigenvalues. Given that each fine-tuned model $\theta_i$ represents a local minimum for its respective task, moving along loss-insensitive directions is helpful for preventing $\theta$ from increasing loss of each task, thereby balancing the fine-tuned models and potentially targeting a low-loss basin shared by all tasks.

As $g(\theta)$ is convex, setting its gradient to zero leads to the closed-form solution:
\begin{equation*} \small
\begin{split}
\frac{\partial g(\theta)}{\partial \theta}  & = 2 \sum_{i=1}^{\mathrm{M}} \left[ \Lambda_{i}^{1/2} Q_{i}^{\top} (\theta - \theta_{i}) \right]^{\top} \frac{\partial \Lambda_{i}^{1/2} Q_{i}^{\top}(\theta - \theta_{i})}{\partial \theta} \\
& = 2\sum_{i=1}^{\mathrm{M}} \left[ \Lambda_{i}^{1/2} Q_{i}^{\top}(\theta - \theta_{i}) \right]^{\top} \Lambda_{i}^{1/2} Q_{i}^{\top} \\
& = 2\sum_{i=1}^{\mathrm{M}} (\theta - \theta_{i})^{\top} F_{\theta_i} = 0
\end{split}
\end{equation*}
which becomes equivalent to Fisher Merging:
\begin{equation} \small
    \theta^{*} = \left( \sum_{i=1}^{\mathrm{M}} F_{\theta_i} \right)^{-1} \left( \sum_{i=1}^{\mathrm{M}} F_{\theta_i} \theta_{i} \right)
\end{equation}

In practice, $F_{\theta_i}$ is replaced by its diagonal approximation to reduce computational complexity \citep{matena2022merging}, which can be seen as assuming independence between the parameters (i.e., $Q_i = I$) \citep{tam2024merging}.

\begin{figure}[t]
    \centering
    \includegraphics[width=1\columnwidth]{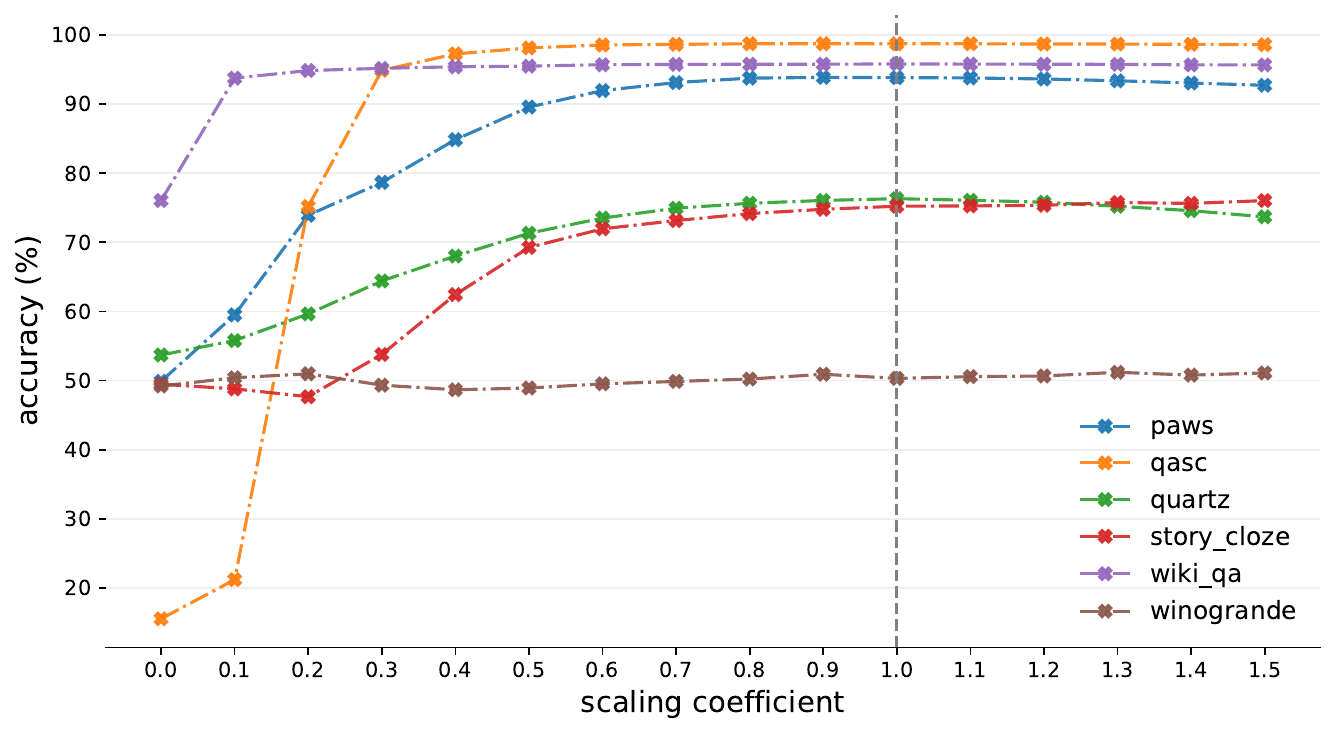}
    \caption{ Accuracy (y-axis) of linear inter/extrapolation between the pre-trained model and fine-tuned models with varying coefficients $\lambda$ (x-axis), T5-base.  }
    \label{fig:pilot_study_scaling}
\end{figure}

\begin{figure*}[tbp]
    \centering
    \includegraphics[width=1.0\textwidth]{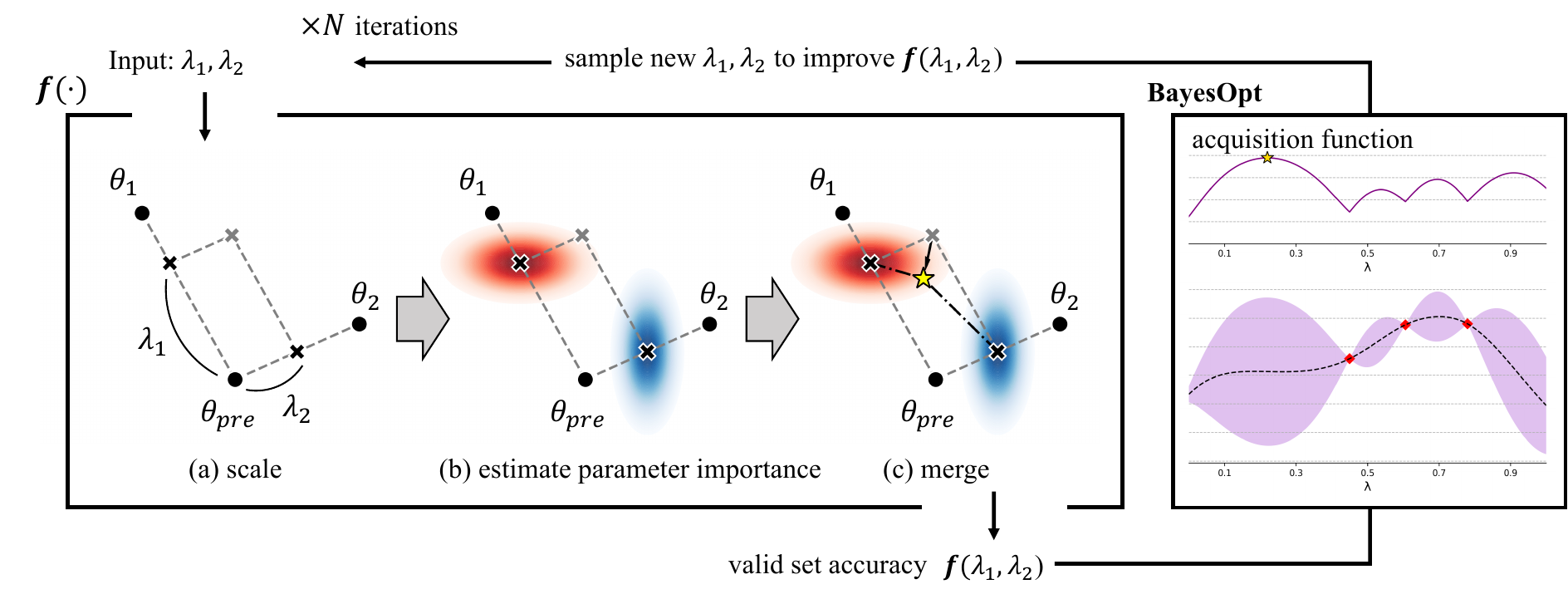}
    \caption{ An illustration of \textbf{DF-Merge}. (1) The black-box function $f(\cdot)$ takes coefficients as inputs to (a) scale the task vectors, and (c) merges models (yellow star) after (b) accounting for the parameter importance using FI, where a contour depicts the local loss landscape of a specific task. (2) The validation set accuracy $f(\cdot)$ is used by Bayesian optimization to suggest the best guess on the coefficients for the next iteration that improve $f(\cdot)$.}
    \label{fig:model}
    %\vspace{-3mm}
\end{figure*}

\section{Dynamic Fisher-weighted Merging}
\label{sec:method}
Performance drop after merging models fine-tuned on different tasks occurs due to parameter interference, in which each task vector may represent a loss-increasing direction for the other tasks. Our pilot study (Figure~\ref{fig:pilot_study_scaling}) shows that linearly interpolating between a pre-trained model and a fine-tuned model reveals numerous alternative local minima, motivating us to search for a set of coefficients such that applying Fisher Merging on the models interpolated by the coefficients minimizes parameter interference. We apply Bayesian optimization for an efficient search of optimal coefficients. An overview of our proposed method, \textbf{DF-Merge}, is illustrated in Figure~\ref{fig:model}. In the following paragraphs, we present an unified view of model merging, then proceed to the details of DF-Merge, divided into merge function and coefficient optimization.

\paragraph{An Unified View of Model Merging.}

We show that Fisher Merging (parameter-wise) and Task Arithmetic (model-wise) both falls under the restricted cases of a more generic form of model merging. This generalized perspective offers a natural way to link both approaches. In particular, we propose a general function of model merging $f(\lambda_1, ..., \lambda_M; \theta_1, ..., \theta_M)$
\begin{equation} \small
\begin{split}
f = \left( \sum_i^M C_{\theta_i} \right)^{-1} \left( M\sum_i^M C_{\theta_i} \cdot \lambda_{i}\tau_{i} \right) + \theta_{pre}
\end{split}
\end{equation}
where $C_{\theta_i}$ is a covariance matrix (e.g., Fisher Information) that depends on $\theta_i$.

This formulation recovers Averaging by setting $\lambda_{i}=1/M$ and $C_{\theta_i}=I$. General Task Arithmetic (GTA) follows from $C_{\theta_i }=I$ while Fisher Merging is obtained with $C_{\theta_i}=\text{diag}(\hat{F}_{\theta_{i}})$ and $\lambda_{i}=1/M$. GTA and Fisher Merging make orthogonal improvements over Averaging: GTA removes the implicit restriction of $\lambda_{i} = 1/M$, whereas Fisher Merging refines parameter importance by replacing $C_{\theta_i }=I$ with $\text{diag}(\hat{F}_{\theta_{i}})$.

Drawing from this insight, we propose the merge function of \textbf{DF-Merge} via linking the benefits of the two:
\begin{multline} \small
    f = \left( \sum_{i}^{M} \text{diag}(\hat{F}_{\theta_{i}(\lambda_{i})})\right)^{-1}\\ \left( M \sum_{i}^{M} \text{diag}(\hat{F}_{\theta_{i}(\lambda_{i})}) \lambda_{i} \tau_{i} \right) + \theta_{pre}
\label{eq:df-merge}
\end{multline}
where $\text{diag}(\hat{F}_{\theta_{i}(\lambda_{i})})$ is the diagonal Fisher Information estimated at $\theta_i(\lambda_i):=\lambda_i \tau_{i} + \theta_{pre}$. Intuitively, this allows Fisher Information to be estimated with varying $\lambda_i$ along the path connecting $\theta_{pre}$ and $\theta_{i}$, unlike Fisher Merging with fixed $\text{diag}(\hat{F}_{\theta_{i}(1)})$.

\paragraph{Coefficient Optimization.}
We employ Bayesian optimization to determine the coefficients $\{\lambda_{i}\}_{i=1}^{M}$ of Eq.\ref{eq:df-merge} that maximize average accuracy on the held-out validation sets. Unlike gradient descent, Bayesian optimization is well-suited for optimizing non-differentiable metrics like accuracy, precision or recall, which directly aligns with the goal of model merging. In addition, Bayesian optimization finds near-optimal coefficients within a few iterations, making it far more scalable than grid search as the number of models increases.

We utilize Gaussian Process to maximize the black box function $f_b(\lambda)$ ($\lambda:=[\lambda_1, ..., \lambda_M] \in \mathbb{R}^{M}$) that returns a scalar metric (i.e., average accuracy) given the merging coefficients $\lambda$. Specifically, the Gaussian process prior is placed over the initial random observations on $t$ points \citep{williams2006gaussian, frazier2018tutorial}:
\begin{equation}
    f_b(\lambda^{1:t}) \sim \mathcal{N}\left(\mu_{0}(\lambda^{1:t}), \Sigma_{0}(\lambda^{1:t},\lambda^{1:t})\right)
\end{equation}
where $\lambda^{1:t}$ is a compact representation of the collection of $t$ points $[\lambda^{1}, ..., \lambda^{t}]$, and $\mu_0$ and $\Sigma_0$ are the mean function and covariance function. Then, the \textit{posterior distribution} of the value of the next point $f_b(\lambda^{t+1})$ is updated by the Bayes' Rule:
\begin{equation}
    f_b(\lambda^{t+1})|f_b(\lambda^{1:t}) \sim \mathcal{N}\left(  \mu_t(\lambda^{t+1}), \sigma_t^2(\lambda^{t+1}) \right)
\end{equation}
where $\mu_t(\lambda^{t+1})$ and $\sigma_t^2(\lambda^{t+1})$ are defined as:
\begin{multline*}
    \mu_t(\lambda^{t+1}) = \Sigma_0(\lambda^{t+1}, \lambda^{1:t})\Sigma_0(\lambda^{1:t}, \lambda^{1:t})^{-1}\\ \cdot(f_b(\lambda^{1:t}) - \mu_0(\lambda^{1:t})) + \mu_0(\lambda^{t+1})
    \\
    \sigma_t^2(\lambda^{t+1}) = \Sigma_0(\lambda^{t+1}, \lambda^{t+1}) - \Sigma_0(\lambda^{t+1}, \lambda^{1:t})\\ \cdot\Sigma_0(\lambda^{1:t}, \lambda^{1:t})^{-1}\Sigma_0(\lambda^{1:t}, \lambda^{t+1}).
\end{multline*}

Subsequently, the next point $\lambda^{t+1}$ to sample is determined by the \textit{acquisition functions}, and we consider Expected Improvement (EI) \citep{frazier2018tutorial} and Upper Confidence Bound (UCB) \citep{ucb2010} in our experiments. EI chooses $\lambda^{t+1}$ such that it maximizes the expected value of improvement than the current best value $f_b^*(t)$ over its posterior distribution:
\begin{equation}
    \mathop{\arg \max}_{\lambda^{t+1}} E_{f_b(\lambda^{t+1})}\left[ \max(f_b(\lambda^{t+1}) - f_b^*(t), 0) \right].
\end{equation}
UCB selects $\lambda^{t+1}$ such that it maximizes the peak of the confidence interval at $\lambda^{t+1}$:
\begin{equation}
    \mathop{\arg \max}_{\lambda^{t+1}} \mu_t(\lambda^{t+1}) + \beta^{1/2} \sigma_t(\lambda^{t+1})
\end{equation}
where $\beta$ is a constant that balances the exploration-exploitation tradeoff. The sampling process is repeated until it reaches the pre-defined number of iterations or the metric converges.

\section{Experiments}

\subsection{Experimental Setup} \label{sec:setup}

\paragraph{Models and Datasets.} We use \textbf{T5-base} and \textbf{T5-large} \citep{raffel2020exploring} which are based on the encoder-decoder architecture and pre-trained on a large-scale corpus with denoising objectives. Both task-specific and multi-task models are fine-tuned on six datasets: PAWS \citep{zhang-etal-2019-paws}, QASC \citep{khot2020qasc}, QuaRTz \citep{tafjord-etal-2019-quartz}, Story Cloze \citep{sharma-etal-2018-tackling}, WikiQA \citep{yang-etal-2015-wikiqa} and Winogrande \citep{sakaguchi2021winogrande}. These datasets cover a range of NLP tasks, including question answering, paraphrase identification, sentence completion, and coreference resolution. See Table~\ref{tab:dataset_statistics} for the dataset statistics\footnote{For datasets without a publicly available labeled test set, the validation set is split into two halves to create new validation and test sets. For datasets with only validation and test sets, the validation set is used for training, and the test set is split into two halves to form new validation and test sets.}. The inputs and outputs are formatted in natural language using the templates in PromptSource \citep{bach2022promptsource} toolkit. See details of training and testing in Appendix~\ref{appendix:training_testing_details}. Note that training is only for simulating model merging under controlled environment, and we posit no access to the training data during merging.

\begin{table}[tbp]
\small
\tabcolsep=0.1cm
\centering
\begin{adjustbox}{width=\columnwidth,center}
    \begin{tabular}{lcccc}
    \toprule
    \textbf{Dataset} & \textbf{\# train} & \textbf{\# validation} & \textbf{\# test} & \textbf{ Task Type} \\
    \midrule
    PAWS & 49401             & 8000                   & 8000        & Paraphrase Identification     \\
    QASC        & 8134              & 463                    & 463     & Question Answering         \\
    QuaRTz       & 2696              & 384                    & 784       & Question Answering       \\
    Story Cloze   & 1871              & 935                    & 936       & Sentence Completion       \\
    WikiQA        & 20360             & 2733                   & 6165       & Question Answering      \\
    Winogrande    & 40398             & 633                    & 634      & Coreference Resolution \\       
    \bottomrule
    \end{tabular}
\end{adjustbox}
\caption{Dataset Statistics.}
\label{tab:dataset_statistics}
\vspace{-3mm}
\end{table}

\paragraph{Evaluation Metric.} All tasks are evaluated by accuracy.

\paragraph{Baselines.}

We compare our approach with several state-of-the-art baselines, including \textbf{Averaging} \cite{wortsman2022model}, \textbf{Fisher Merging} \cite{matena2022merging}, \textbf{Task Arithmetic} \cite{ilharco2023editing}, \textbf{DARE} \cite{yu2024language} and \textbf{TIES-Merging} \cite{yadav2024ties}. TIES-Merging employs trim, elect, and disjoint mean operation to resolve parameter interference between the fine-tuned models. DARE randomly drops and re-scales the task vectors to sparsify them, potentially alleviating the parameter interference.

\paragraph{Implementation Details.} 
\textbf{DF-Merge:} We optimize the coefficients using Bayesian Optimization package \citep{bayesopttoolkit} to maximize average accuracy on held-out validation sets. DF-Merge runs for $50$ iterations, preceded by 10 random initialization steps, with coefficients constrained to the range of $[0, 1]$. For each iteration, \( \text{diag}(\hat{F}) \) is computed exactly over the model's predictive distribution using $30$ unlabeled validation samples. \textbf{Baselines:} The best coefficients of Task Arithmetic and TIES-Merging are determined by a grid search (TA: $[0, 1]$, TIES: $[0.8, 1.8]$) on validation sets with a step size of $0.1$. DARE is applied on TA, with additional grid search over the drop rate $p$ in $[0.1, 0.9]$ with a step size of $0.2$. Unless otherwise stated, experimental results are averaged over five random runs with significance testing.

\begin{table*}[t]
    \small
    \centering
    \begin{adjustbox}{width=\textwidth,center}
    \begin{tabular}{cc|c|ccccccc}
    \toprule
    \textbf{Model} & \textbf{Method} & \parbox[t]{1.5cm}{\centering \textbf{Valid. Set}} & \textbf{PAWS} & \textbf{QASC} & \textbf{QuaRTz} & \textbf{Story Cloze} & \textbf{WikiQA} & \textbf{Winogrande} & \textbf{Avg.} \\
    \midrule
    \multirow{10.5}{*}{T5-base}
    & Zero-shot & - & 49.89 & 15.55 & 53.70 & 49.47 & 76.04 & 49.21 & 48.98 \\
    & Fine-tune & - & 93.81 & 98.70 & 76.30 & 75.21 & 95.79 & 50.32 & 81.69 \\
    & Multi-task & - & 93.26 & 98.49 & 66.38 & 80.73 & 95.46 & 56.09 & 81.73 \\
    \cmidrule{2-10}
    & Averaging & \xmark  & 68.92 & 82.33 & 59.72 & 49.74 & 94.30 & \underline{50.79} & 67.63* \\
    & Fisher Merging & \cmark & 88.47 & 84.02 & 64.64 & 52.76 & 94.79 & \textbf{51.04} & 72.62* \\
    & Task Arithmetic & \cmark & 79.75 & 88.21 & 62.81 & 67.09 & \underline{95.17} & 48.64 & 73.61* \\
    & DARE & \cmark &  79.80 & 87.82 & 62.86 & 67.35 & 95.16 & 48.96 & 73.66* \\
    & TIES-Merging & \cmark & \textbf{90.18} & 78.32 & 61.53 & 57.78 & 95.10 & 49.43 & 72.06* \\
    \cmidrule{2-10}
    & DF-Merge~(EI) & \cmark & \underline{89.62} & \textbf{97.37} & \textbf{68.21} & \textbf{68.97} & 95.02 & 49.62 & \textbf{78.14} \\
    & DF-Merge~(UCB) & \cmark & 89.58 & \underline{96.11} & \underline{66.63} & \underline{68.70} & \textbf{95.19} & 49.21 & \underline{77.57} \\ 
    \midrule
    \midrule
    \multirow{10.5}{*}{T5-large}
    & Zero-shot & - & 55.39 & 11.23 & 54.97 & 50.32 & 70.79 & 48.42 & 48.52 \\
    & Fine-tune & - & 94.36 & 98.32 & 86.43 & 90.77 & 96.16 & 54.42 & 86.74 \\
    & Multi-task & - & 94.29 & 99.18 & 83.11 & 89.83 & 95.94 & 67.54 & 88.31 \\
    \cmidrule{2-10}
    & Averaging & \xmark & 75.27 & 35.08 & 70.64 & 57.22 & 86.83 & 50.00 & 62.51* \\
    & Fisher Merging & \cmark & 67.70 & 61.64 & \underline{81.30} & 68.35 & 89.16 & 51.10 & 69.88* \\
    & Task Arithmetic & \cmark & 90.86 & 95.46 & 73.05 & 84.78 & 93.24 & \textbf{53.79} & 81.86\\
    & DARE & \cmark & \underline{91.01} & 95.68 & 72.07 & 84.66 & 93.20 & 52.84 & 81.58* \\
    & TIES-Merging & \cmark & \textbf{93.12} & 93.05 & 69.21 & 79.91 & 92.51 & \underline{53.60} & 80.23* \\
    \cmidrule{2-10}
    & DF-Merge~(EI) & \cmark & 89.43 & \underline{96.46} & \underline{81.30} & \textbf{86.99} & \textbf{95.15} & 52.24 & \textbf{83.59} \\
    & DF-Merge~(UCB) & \cmark & 89.94 & \textbf{96.76} & \textbf{81.68} & \underline{85.79} & \underline{94.70 } & 51.86 & \underline{83.46} \\
    \bottomrule
    \end{tabular}
    \end{adjustbox}
    \caption{Evaluation result (\%) of DF-Merge and the baselines on six tasks. The best accuracy is bolded and the second-best accuracy is underlined, for each column of a type of model. *: Both DF-Merge (EI) and DF-Merge (UCB) significantly outperform the baseline ($p<0.05$).}
    \label{tb:main_results}
\end{table*}

\subsection{Main Results}\label{sec:main_res}

Table~\ref{tb:main_results} shows the performance of DF-Merge and the baselines. There are several key observations from the results. \textbf{First,} DF-Merge outperforms the baselines in average accuracy by large margins and the improvements are significant for almost all baselines. In particular, DF-Merge improves over the best baseline in the average accuracy by 4.48 point for T5-base and 1.73 point for T5-large. \textbf{Second,} DF-Merge narrows the gap with the oracle multi-task learning model by a substantial degree. For example, the gap in average accuracy can be narrowed down to 3.55 point for T5-base and 3.15 point for T5-large. This result indicates that DF-Merge can be a useful training-free alternative to multi-task learning in settings where a slight performance drop is permissible. \textbf{Third,} DF-Merge strikes an adequate balance between the performances across multiple tasks. For instance, the maximum drop in accuracy compared to the fine-tuned model among the six tasks is 8.09 point for T5-base and 4.75 point for T5-large, which are smaller than all baselines. In contrast, the baselines tend to build a multi-task model that excels in one task yet at the cost of compromising the performance of other tasks. For instance, Fisher Merging achieves a notable accuracy on QuaRTz (T5-large) while being much worse on the remaining tasks than the other methods. 

\section{Analysis and Discussion}

\begin{table}[t]
\small
\tabcolsep=0.1cm
\centering
\begin{adjustbox}{width=0.9\columnwidth,center}
  \begin{tabular}{lcccc}
    \toprule
    \bf Method & \bf T5-base  & \bf T5-large \\
    \midrule
    \textbf{DF-Merge~(EI)} & 78.14 & 83.59 \\
    \enspace w/o Fisher Information & 76.80*~(-1.34) & 82.52~(-1.07) \\
    \enspace w/o Bayesian Optimization & 72.62*~(-5.52) & 69.88*~(-13.71) \\
    \midrule
    \textbf{DF-Merge~(UCB)} & 77.57 & 83.46 \\
    \enspace w/o Fisher Information & 76.72*~(-0.85) & 82.49~(-0.97) \\
    \enspace w/o Bayesian Optimization & 72.62*~(-4.90) & 69.88*~(-13.58) \\
    \midrule
    \textbf{Averaging} & 67.63 & 62.51 \\
    \bottomrule
\end{tabular}
\end{adjustbox}
\caption{Ablation of DF-Merge components, evaluated by the average test set accuracy (\%). *: significant drop in performance after ablation ($p<0.05$). }
\label{tab:ablation}
\vspace{-3mm}
\end{table}

\begin{figure*}[t]
    \centering
    \includegraphics[width=\textwidth]{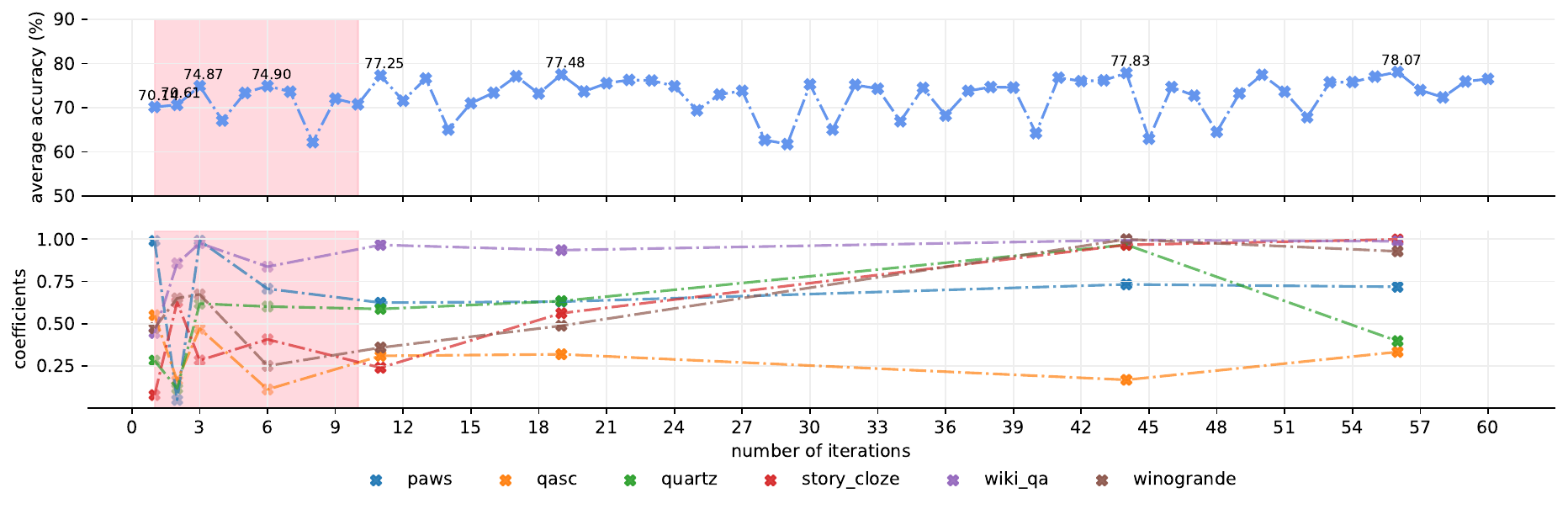}
    \vspace{-8mm}
    \caption{Bayesian optimization trajectory of DF-Merge (T5-base, UCB).  Coefficients (\textbf{bottom}) and the average validation set accuracy (\textbf{top}) are rendered as a function of iterations. The coefficients with the new highest average accuracy up to their corresponding iteration are shown. The red area denotes initial random evaluations.}
    \label{fig:optim_trajectory}
    \vspace{-3mm}
\end{figure*}

\subsection{Ablation Study}

We conduct an ablation study of DF-Merge components to understand each component's contribution to the final performance, as shown in Table~\ref{tab:ablation}. Removing Fisher Information from DF-Merge—i.e., using GTA and selecting coefficients via Bayesian optimization—results in a consistent drop in performance across different model sizes and acquisition functions. Notably, the performance drop is significant with EI, highlighting the importance of leveraging useful information from local loss curvature. Besides, removing Bayesian optimization from DF-Merge—i.e., Fisher Merging—causes significant drops in performance, indicating that DF-Merge benefits largely from a flexible coefficient search. To summarize, both Fisher information and Bayesian optimization are essential to the optimal performance of DF-Merge.

\subsection{Efficiency Analysis}

Though DF-Merge effectively identifies optimal task vector coefficients in a vast search space, it still requires a number of merge-then-evaluate rounds with labeled validation sets, posing a challenge in terms of both computational budget and data labeling cost. Hence we examine whether the effectiveness of DF-Merge remains solid within a few number of iterations as well as with validations sets of reduced sizes. Results in this section are based on a single run with a fixed random seed.

\paragraph{Effect of the number of iterations.}
Figure~\ref{fig:optim_trajectory} demonstrates that DF-Merge quickly achieves the near-optimal performance within a few number of iterations. In particular, after the initial evaluations on 10 random points (red area), it takes 9 iterations for DF-Merge to exploit previous observations and discover near-optimal coefficients with 0.59\%p gap compared to the best ones (56th iteration). 

The remaining iterations are responsible for a marginal improvement, indicating that the major enhancement in performance occurs in the early stage of optimization. We observe similar trends consistently when using EI as the acquisition function or using T5-large, shown in Appendix~\ref{appendix:optimization_trajectories}. Hence the optimization may be terminated early to save much of the runtime and computational resources.

\begin{figure}[t]
    \centering
    \includegraphics[width=\columnwidth]{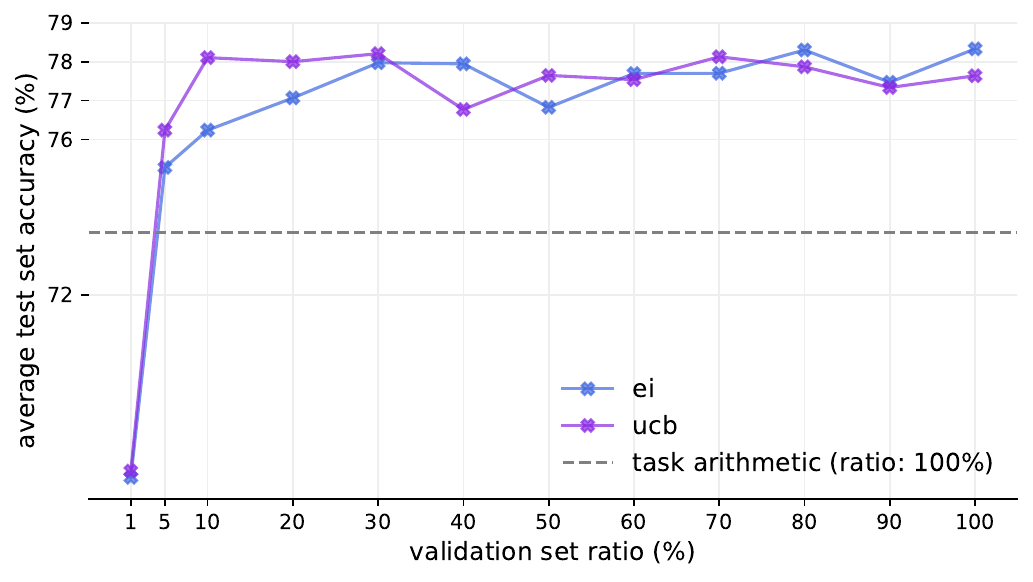}
    \vspace{-8mm}
    \caption{ Average test set accuracy (\%) of DF-Merge with varying ratios of validation samples used, T5-base, based on a single run. }
    \label{fig:val_ratio}
    \vspace{-3mm}
\end{figure}

\paragraph{Effect of the validation set size.} 
We randomly sample varying ratios of data from the validation set of each task with larger sampled sets containing the smaller ones, and observe the \textit{test} set performance of DF-merge, as shown in Figure~\ref{fig:val_ratio}. Similar to the findings for the number of iterations, DF-merge efficiently achieves optimal performance with a minimal size of validation set. Notably, 5\% of the validation data suffices to closely approach the performance of utilizing the full validation sets, as well as to outperform Task Arithmetic by a large margin. Additionally, this trend consistently holds regardless of which acquisition function (EI or UCB) is used when $ratio \geq 30\%$. Consequently, DF-Merge can save the data labeling cost and reduce the computations for running inference on the validation sets while keeping its performance intact. 

\subsection{Metric Landscape of DF-Merge}

\begin{figure}[t]
    \centering
    \includegraphics[width=\columnwidth]{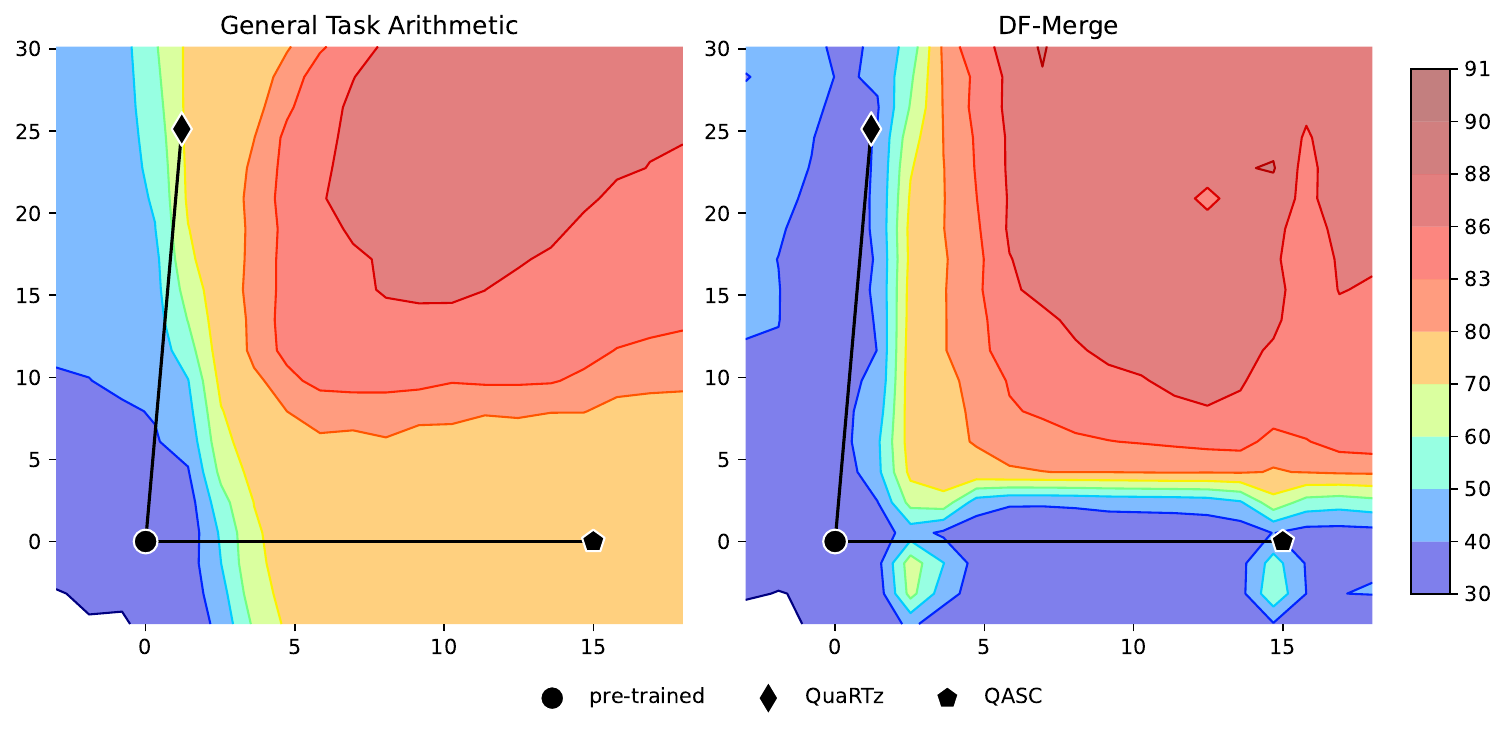}
    \caption{A landscape of average validation accuracy (denoted with colors) of merging two T5-base models, based on a single run. A point in the 2-D plane represents a linear combination of the two task vectors. \textbf{left}: GTA; \textbf{right}: DF-Merge.}
    \label{fig:acc_landscape}
    \vspace{-3mm}
\end{figure}

We examine how DF-Merge improves multi-task merging by analyzing the metric landscape of General Task Arithmetic and DF-Merge. In particular, we visualize instances of GTA and DF-Merge specified by two merging coefficients in a two-dimensional subspace\footnote{Following \citet{garipov2018loss}, we let \(u = \tau_1\), \(\hat{u} = u / \|u\|\), \(v = \tau_2 - \langle \tau_2, \hat{u} \rangle \hat{u}\), and \(\hat{v} = v / \|v\|\). Then, \(\hat{u}\) and \(\hat{v}\) form an orthonormal basis of a 2-D plane, where a coordinate \((\lambda_1, \lambda_2)\) specifies a point in the plane as \(P(\lambda_1, \lambda_2) = \theta_{\text{pre}} + \lambda_1 \hat{u} + \lambda_2 \hat{v}\).}, as shown in Figure~\ref{fig:acc_landscape}. 

We make the following key observations. \textbf{First,} the high accuracy region resides in the upper-right side of the landscape, indicating that the optimal accuracy is unlikely to appear when all coefficients are low. \textbf{Second,} GTA and DF-Merge both have a large and flexible search space of the coefficients, compared to TA and Averaging. Overall, DF-Merge has a broader high-accuracy regions compared to GTA, possibly since DF-Merge incorporates \textit{parameter-wise} importance given by Fisher Information. Based on the analysis in Section~\ref{sec:background}, using Fisher Information can also be interpreted as merging along the low-loss basin. \textbf{Third,} DF-Merge underperforms GTA when the coefficients are low, as seen in the lower-right and upper-left side of the landscape. A possible reason is that a model scaled with a low coefficient no longer remains as a local minimum to guarantee the low-loss preserving property of Fisher information.

\section{Related Work}
\paragraph{Foundations of Model Merging.} 
Recent studies have shown that models sharing the same initialization reside in the same low-loss basin, often connected by a path with non-increasing loss, known as \textit{mode connectivity} \citep{garipov2018loss, draxler2018essentially, mirzadeh2021linear}. On the contrary, barriers often exists between models optimized from different initialization \citep{neyshabur2020being}. \citet{entezari2022the} shows that SGD solutions from different random initialization can be teleported to the same low-loss basin after accounting for the permutation invariance of neural network. This idea has been introduced to merging models with different initializations \citep{ainsworth2023git, stoica2024zipit}. In this paper, we focus on merging fine-tuned models from the same pre-trained initialization.

\paragraph{Building Multitask Model via Merging.} An important application of model merging is building a multitask model out of multiple task-specific models fine-tuned from the same initialization (e.g., pre-trained model). While simple averaging is a strong baseline that improves single task merging \citep{wortsman2022model}, it falls significantly short when applied to multitask scenarios. This has led to a series of methods which bridge the gap with multitask fine-tuned models: Fisher Merging \citep{matena2022merging} frames merging as a maximization of joint posterior of models' parameters. RegMean \citep{jin2023dataless} minimizes regression errors between the merged model and the fine-tuned models. Unlike previous methods aiming to find closed-form solution, \citet{tam2024merging} shows that their iterative method can solve an improved merging objective which is intractable to solve analytically. Meanwhile, Task Arithmetic (TA) \citep{ilharco2023editing} presents a scalable approach for editing fine-tuned parameters to guide the behavior of pre-trained models, a theoretical analysis of which suggests that weight disentanglement arising from pre-training is what makes TA successful \citep{ortiz-jimenez2023task}. Building on these findings, recent works has explored effective methods for editing fine-tuned parameters with different emphasis, such as resolving parameter interference \citep{yadav2024ties, daheim2024model}, sparsifying task vectors \citep{yu2024language,  davari2024modelbread, deep2024della}, training coefficients \citep{yang2024adamerging}, or applying to adapters \citep{tang2024parameterefficient}.

\paragraph{Bayesian Optimization in NLP.} Bayesian optimization is a family of iterative algorithms for efficient hyperparameter search over a black-box function that is expensive to evaluate. Its applications are found in a range of tasks in NLP, such as optimizing hyperparmeters for text representation \citep{yogatama2015bayesian}, data selection criteria \citep{ruder2017learning}, and model ensemble \citep{pour2024gaussian}. Most importantly, \citet{liu2024checkpoint} utilize Bayesian optimization to find coefficients for average merging that improve checkpoint merging during LLM pre-training. Instead, we leverage Bayesian optimization conditioned on our newly proposed merging objective.

\section{Conclusion}

In this work, we propose an unified merging framework and introduce Dynamic Fisher-weighted Merging (\textbf{DF-Merge}). This approach assigns scaling coefficients to fine-tuned model parameters and dynamically adjusts them using Bayesian optimization, with the goal of maximizing validation performance. Through this process, DF-Merge tries to efficiently identify low-loss basins using Fisher information. Experimental results demonstrate that DF-Merge consistently outperforms strong baselines across models of different sizes on diverse tasks. The method proves effective in achieving near-optimal performance in just a few iterations, even with minimal validation data, highlighting its potential as a powerful tool for multitask model merging.

\section*{Limitations}
In this section, we discuss the limitations of our work as follows. 
\textbf{First}, DF-Merge requires the fine-tuned model share the same architecture and pre-trained parameters. Though DF-Merge covers a majority of merging settings given the prevalence of fine-tuning the same pre-trained model, there indeed exist scenarios where one wish to fuse the distinct task expertise of models with different initializations or even across incompatible architectures. We leave this direction for the future research. \textbf{Second}, DF-Merge relies on the labeled validation sets, albeit with a relatively small number of samples required to achieve optimal performance. Yet we believe there may be ways to apply DF-Merge when the validation sets are not available. For instance, the fine-tuned models could serve as the pseudo-labeler at the test time, in which case the merging objective becomes maximally replicating each model's predictions on the test inputs. Since we do not leverage the label when estimating the Fisher information, the above approach is feasible. \textbf{Third}, our use of Fisher Information is restricted to its diagonal simplification, as the Fisher Information is intractable to compute given its extremely large number of entries ($O(d^2)$ with $d$ being the number of model parameters) for modern PLMs. Diagonal Fisher information implicitly supposes the model parameters are not related to each other in terms of gradient, which is a strong assumption that might lead to suboptimal performance. A promising research direction would be relaxing this assumption, such as representing Fisher Information as a block-diagonal matrix \citep{tam2024merging}.

\section*{Ethics Statement}
\paragraph{Potential Risks}
If some of the fine-tuned models are trained for malicious purpose, then the merged model DF-Merge might risk producing biased predictions, harmful contents, or unfair decisions, even if the safety of other models are guaranteed. Our method does not address these potential risks, therefore the safety of the merged mode must be checked before deployment. 

\paragraph{Use of Scientific Artifacts}
The \textbf{Bayesian Optimization} \citep{bayesopttoolkit} is under MIT license and \textbf{PromptSource} \citep{bach2022promptsource} is under Apache-2.0 license, both of which permits the use of the tool for research purpose. For the datasets used in our experiments, \textbf{PAWS} \citep{zhang-etal-2019-paws} permits its free use for any purpose, QASC \citep{khot2020qasc} is under the CC BY 4.0 license, QuaRTz \citep{tafjord-etal-2019-quartz} and Story Cloze \citep{sharma-etal-2018-tackling} are under Creative Commons License, Winogrande \citep{sakaguchi2021winogrande} is under Apache, and WikiQA {\citep{yang-etal-2015-wikiqa}} is licensed under Microsoft Research Data License Agreement for Microsoft Research WikiQA Corpus. These datasets are publicly available for research purpose. The datasets are intended to serve as benchmarks for testing the ability of AI models on language tasks, hence our experiments are aligned with the intended use.

\paragraph{Model Size and Computational Budget}
We use T5-base and T5-large \citep{raffel2020exploring} which have 223 million and 738 million parameters, respectively. DF-Merge is cost-effective compared to training models, where running a single iteration of DF-Merge approximately requires 70 seconds for T5-base and 170 seconds for T5-large on a single A100 GPU.
%\clearpage

\bibliography{custom}

\appendix
\section{Training \& Testing  Details of T5 Models}
\label{appendix:training_testing_details}
Training is conducted with a batch size of 64, using the AdamW \citep{loshchilov2018decoupled} optimizer, a fixed learning rate of \(1 \times 10^{-4}\), and 2,500 steps for each task-specific model, while the multitask model is trained for 25,000 steps with early stopping. The model with the lowest validation loss is selected for testing. For each test input, we forward the input/output pairs of all possible labels to the model and select the one with the lowest perplexity as the final prediction.

\section{Optimization Trajectories}
\label{appendix:optimization_trajectories}

\begin{figure*}[htbp]
    \centering
    \includegraphics[width=\textwidth]{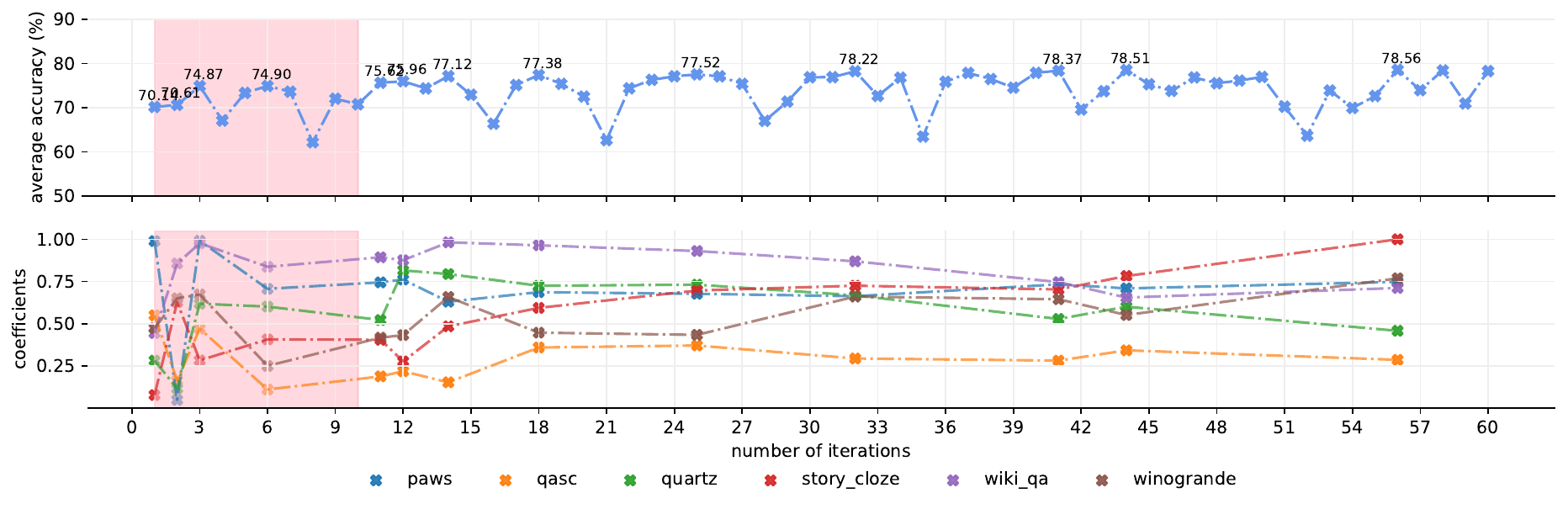}
    \caption{Bayesian optimization trajectory of DF-Merge (T5-base, EI).}
    \label{fig:trajectory-t5-base-ei-grid}
\end{figure*}

\begin{figure*}[htbp]
    \centering
    \includegraphics[width=\textwidth]{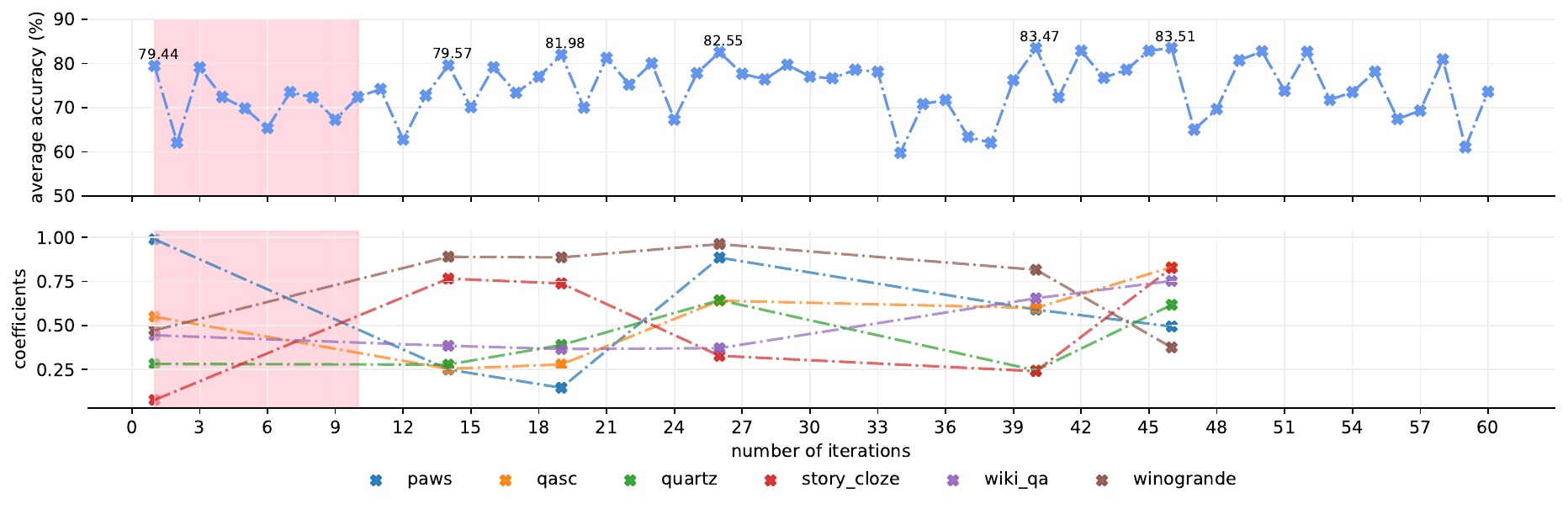}
    \caption{Bayesian optimization trajectory of DF-Merge (T5-large, EI).}
    \label{fig:trajectory-t5-large-ei-grid}
\end{figure*}

\begin{figure*}[htbp]
    \centering
    \includegraphics[width=\textwidth]{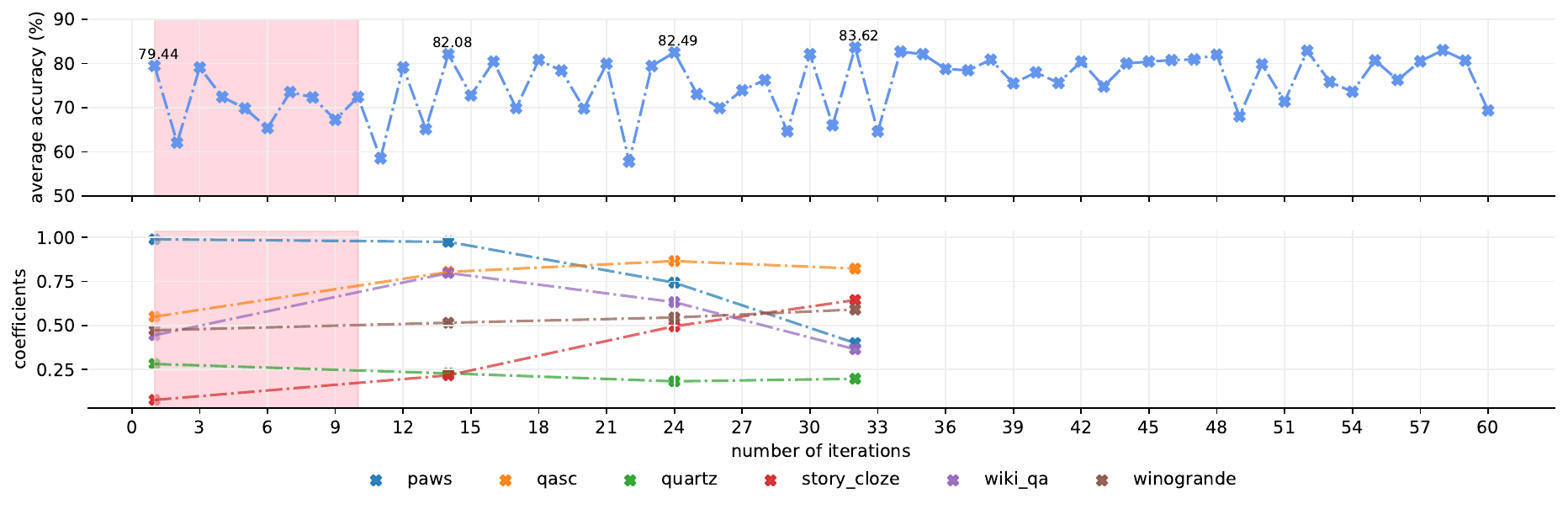}
    \caption{Bayesian optimization trajectory of DF-Merge (T5-large, UCB).}
    \label{fig:trajectory-t5-base-ucb-grid}
\end{figure*}

We complement the optimization trajectories of DF-Merge for T5-base + EI, T5-base + UCB and T5-large + EI in Figure~\ref{fig:trajectory-t5-base-ei-grid}, Figure~\ref{fig:trajectory-t5-base-ucb-grid} and Figure~\ref{fig:trajectory-t5-large-ei-grid}, respectively.

\end{document}